\documentclass[10pt]{article} 

\usepackage[preprint]{rlj} 

\usepackage{amssymb}            
\usepackage{mathtools}          
\usepackage{mathrsfs}           
\usepackage{graphicx}           
\usepackage{subcaption}         
\usepackage[space]{grffile}     
\usepackage{url}                
\usepackage{lipsum}             
\usepackage{booktabs}
\usepackage{wrapfig}
\usepackage[dvipsnames,table]{xcolor}

\NewDocumentCommand{\HHRL}{}{H$^2$RL}
\NewDocumentCommand{\hhrlp}{}{H$^2$RL\textsubscript{+}}
\NewDocumentCommand{\hhrlpp}{}{H$^2$RL\textsubscript{++}}

\newcommand{\spm}[1]{\textsubscript{$\pm$#1}}

\title{ Boosting Deep Reinforcement Learning using\\ Pretraining with Logical Options}

\setrunningtitle{Boosting deep Reinforcement Learning using pretraining with Logical Options}

\author{Zihan Ye\textsuperscript{1,2,5}, Phil Chau\textsuperscript{1,2}, Raban Emunds\textsuperscript{1,2,5},\\Jannis Blüml\textsuperscript{1,2,5}, Cedric Derstroff\textsuperscript{1,2,5}, Quentin Delfosse\textsuperscript{2},\\ Oleg Arenz\textsuperscript{1,3}, Kristian Kersting\textsuperscript{1,2,4,5}}

\emails{zihan.ye@tu-darmstadt.de}

\affiliations{
$^{1}$\textbf{CS Department, Technical University of Darmstadt, Germany}\\
$^{2}$\textbf{AIML Group}
$^{3}$\textbf{IAS Group}
$^{4}$\textbf{DFKI}
$^{5}$\textbf{Hessian AI}
\par 
}

\contribution{
We introduce \HHRL{}, a hierarchical neuro-symbolic reinforcement learning (RL) framework designed to mitigate policy misalignment in deep RL. Leveraging logic-informed pretraining, \HHRL{} embeds logic priors directly into neural policies, facilitating task-semantic aligned decision-making while eliminating the computational overhead of logic reasoning and symbol extraction at inference time.
    }
    {
    Deep RL policies are frequently misaligned: they often achieve high returns based on shortcut behaviors, or reward-hacking strategies that deviate from the intended task semantics
    ~\citep{Geirhos20Shortcuts, koch21objective, Hermann24shortcut}. Prior neuro-symbolic RL methods, e.g., BlendRL~\citep{shindoblendrl}, NUDGE~\citep{delfosse2023interpretable}, and NLRL~\citep{jiang2019neural} achieve policy alignment, but employ their logical policies directly on low-level actions and require online logic reasoning during inference time, introducing additional computational overhead and unnecessary dependence. 
    \HHRL{}, to the best of the authors' knowledge, is the first pretraining framework that leverages logic-informed pretraining to embed logic priors directly into neural RL policies, mitigating policy misalignment in deep RL while preserving fast, purely neural inference at deployment.  
    }

\contribution{
We conduct ablation studies, which demonstrate that logic-informed pretraining is crucial for mitigating policy misalignment in deep RL.
    }
    {
    While two-stage training remains under-explored in deep reinforcement learning, \HHRL{} addresses this gap by first aligning the neural policy with game semantics through logic-informed pretraining. The neural policy is then subsequently refined via direct environment interaction.
    }

\contribution{
   We demonstrate the versatility of our approach by confirming \HHRL{}'s effectiveness as a general pretraining framework for a range of deep RL algorithms. 
    }
    {
    We verify our approach on DQN~\citep{mnih2015human}, PPO~\citep{schulman2017proximal}, and C51~\citep{c51} policies, which tend to be misaligned in long-horizon tasks (e.g., Kangaroo; cf. Contr. 1.). \HHRL{}-pretraining successfully addresses the policy misalignment issue as shown in Sec.~\ref{experiments}.
    }

\keywords{Neuro-symbolic RL, Differentiable Reasoning, Deep RL policy pretraining.} 

\summary{Deep reinforcement learning agents are often misaligned, as they over-exploit early reward signals. Recently, several symbolic approaches have addressed these challenges by encoding sparse objectives along with aligned plans. However, purely symbolic architectures are complex to scale and difficult to apply to continuous settings. Hence, we propose a hybrid approach, inspired by humans' ability to acquire new skills. We use a two-stage framework that injects symbolic structure into neural-based reinforcement learning agents without sacrificing the expressivity of deep policies. Our method, called \textit{Hybrid Hierarchical RL (\HHRL)}, introduces a logical option-based pretraining strategy to steer the learning policy away from short-term reward loops and toward goal-directed behavior while allowing the final policy to be refined via standard environment interaction. Empirically, we show that this approach consistently improves long-horizon decision-making and yields agents that outperform strong neural, symbolic, and neuro-symbolic baselines.
}

\begin{document}

\maketitle  

\begin{abstract}
Deep reinforcement learning agents are often misaligned, as they over-exploit early reward signals. Recently, several symbolic approaches have addressed these challenges by encoding sparse objectives along with aligned plans. However, purely symbolic architectures are complex to scale and difficult to apply to continuous settings. Hence, we propose a hybrid approach, inspired by humans' ability to acquire new skills. We use a two-stage framework that injects symbolic structure into neural-based reinforcement learning agents without sacrificing the expressivity of deep policies. Our method, called \textit{Hybrid Hierarchical RL (\HHRL)}, introduces a logical option-based pretraining strategy to steer the learning policy away from short-term reward loops and toward goal-directed behavior while allowing the final policy to be refined via standard environment interaction. Empirically, we show that this approach consistently improves long-horizon decision-making and yields agents that outperform strong neural, symbolic, and neuro-symbolic baselines.
\end{abstract}

\section{Introduction}

\begin{wrapfigure}[20]{t}{0.5\textwidth}
    \centering
    \includegraphics[width=0.5\textwidth]{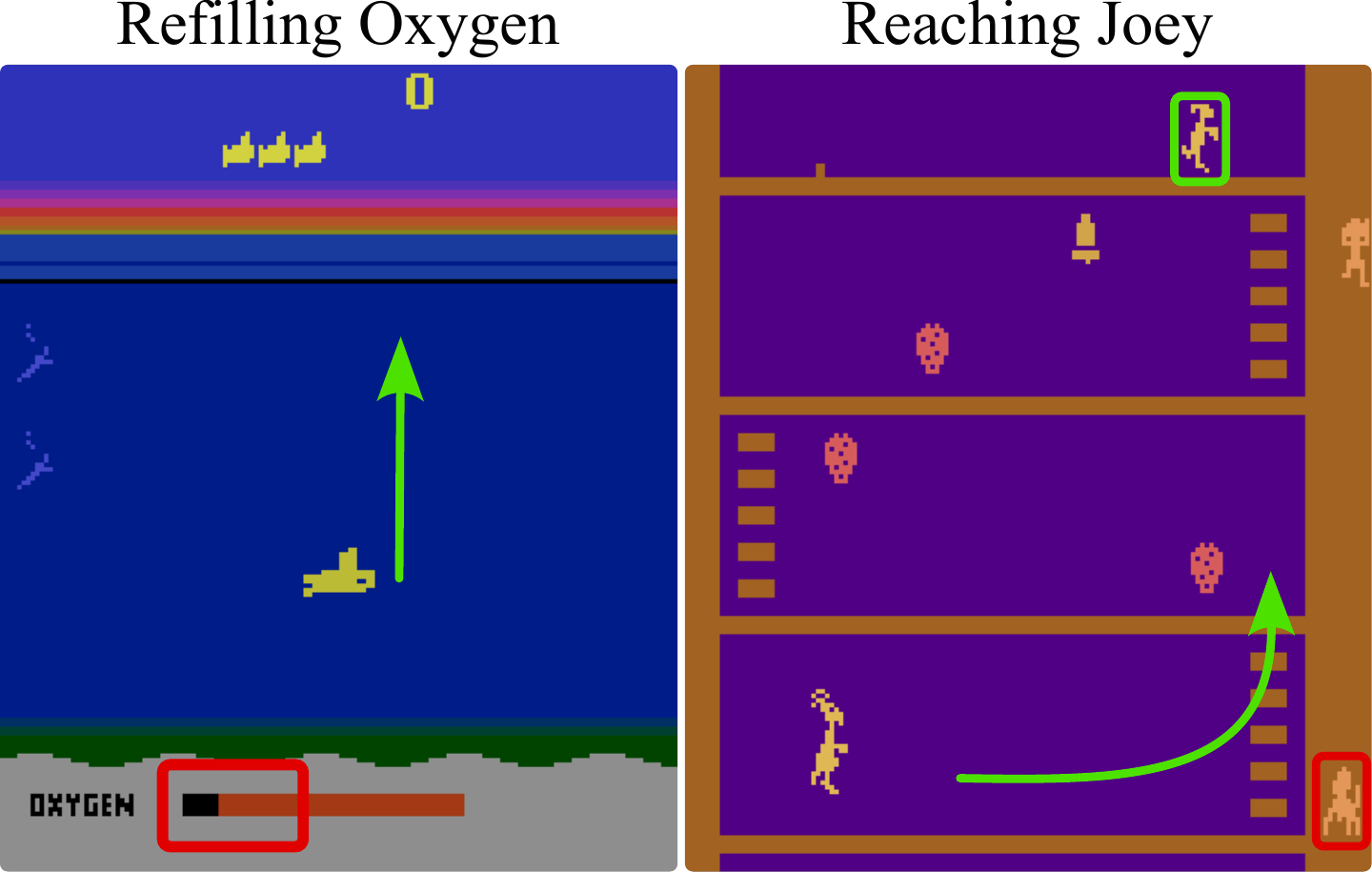}
    \caption{\textbf{Deep reinforcement learning policies are often misaligned}, exemplified on neural PPO agents. Although the oxygen is running low in \textit{Seaquest} (left) and the goal in \textit{Kangaroo} (right) is to go up, PPO agent fails to choose the optimal actions (in green). Instead, they focus on immediate rewards, e.g., keep attacking enemies. \label{fig:ppofail}}
\end{wrapfigure}

In deep reinforcement learning (RL), while sparse rewards hinder exploration, dense rewards designed to guide agents are prone to \textit{reward hacking}~\citep{Everitt21tampering, skalse2022defining, Delfosse25hackatari}. 
As noted by~\citet{Goodhart1984}, ``When a measure becomes a target, it ceases to be a good measure.'' 
This manifests as \textit{shortcut learning}, where agents exploit spurious correlations rather than solving the intended task~\citep{Ilyas19adversarial, Chan20reliable, Geirhos20Shortcuts, koch21objective, Hermann24shortcut, Delfosse25hackatari}. 
For instance, in multi-objective Atari games like \textit{Seaquest} and \textit{Kangaroo} (depicted in Fig.~\ref{fig:ppofail}), deep RL agents often prioritize short-term and easily accessible gains. 
These agents focus on shooting and punching enemies until oxygen or time runs out—over essential long-horizon objectives like refilling oxygen and collecting divers (in \textit{Seaquest}) or climbing ladders to reach the joey (in \textit{Kangaroo}), ultimately leading to failure~\citep{ Delfosse25hackatari}.

To mitigate such pitfalls, many recent approaches have integrated symbolic representations to provide reasoning capabilities similar to human planning~\citep{ye2022differentiable, dylan, delfosse2023interpretable, Grandien24interpretable, Luo24insight, Bhuyan24nesysurvey, delvecchio25nesysurvey, shindoblendrl, nexus}. However, relying on explicit symbolic reasoning during inference creates significant computational overhead and latency, limiting real-time applicability~\citep{Bhuyan24nesysurvey, delvecchio25nesysurvey}. Moreover, due to its discrete nature, it is tricky, if not impossible, to use symbolic reasoning without extra work in a continuous action space. Alternatively, manual reward shaping can guide exploration~\citep{ng99shaping, Gupta22rewardshaping}, but lacks the precision of symbolic logic and requires tedious, domain-specific tuning that is difficult to generalize.

Our approach addresses these limitations by drawing on the cognitive process of \textit{scaffolding} in human learning. Humans rarely learn through unstructured trial-and-error; they rely on explicit instruction and rules to establish fundamentals before transitioning to a phase of ``free play'' to achieve mastery~\citep{bransford2000people, dreyfus1980five}. For example, consider mastering a complex motor skill, such as playing tennis: novices do not begin by learning from playing competitive matches. Rather, they start off by mastering individual components, such as the grip, the swing, and footwork. Only after these foundational mechanics are internalized does the learner progress to the open-ended free play to improve their skills. This suggests that effective learning requires an initial phase of structured guidance to establish a behavioral prior, followed by unconstrained optimization.

Specifically, based on this intuition, we introduce \textit{Hybrid Hierarchical Reinforcement Learning} (\HHRL{}). The core innovation of \HHRL{} is the use of differentiable symbolic logic and options solely during the pretraining phase. This implicitly encodes high-level reasoning and inductive biases into the neural network's parameters, allowing the agent to internalize long-horizon dependencies. Critically, because the symbolic engine is not required in post-training, the final agent retains the inference speed of a standard neural policy while exhibiting the structural coherence of a symbolic reasoner. Empirically, \HHRL{} significantly outperforms baselines in environments with deceptive reward functions. We demonstrate that our agents avoid common misalignment pitfalls and escape reward traps that trap purely neural approaches, effectively solving the trade-off between symbolic control and neural scalability. 

In summary, our contributions are: 
\begin{description}[leftmargin=20pt, itemsep=0pt,parsep=0pt,topsep=-3pt,partopsep=0pt] \item[(i)] 
We introduce \HHRL{}, a hierarchical neuro-symbolic reinforcement learning (RL) framework designed to mitigate policy misalignment in deep RL. Leveraging logic-informed pretraining, \HHRL{} embeds logic priors directly into neural policies, facilitating task-semantic aligned decision-making while eliminating the computational overhead of logic reasoning and symbol extraction at inference time.
\item[(ii)] We conducted ablation studies, demonstrating that logic-informed pretraining is crucial for mitigating policy misalignment in deep RL. 
\item[(iii)] We demonstrate the versatility of our approach by confirming \HHRL{}'s effectiveness as a general pretraining framework for a range of deep RL algorithms. 
\end{description}

We proceed as follows. We start off by discussing background in Sec.~\ref{background}, followed by a detailed introduction of \HHRL{} in Sec.~\ref{HHRL}. Before concluding, we present our experimental results and discuss related work in Secs.~\ref {experiments} and \ref{related work}, respectively.
\begin{figure*}[t]
    \centering
    \includegraphics[width=\textwidth]{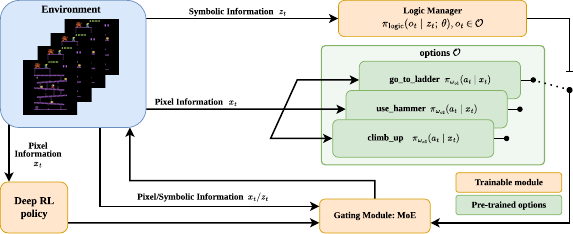}
    \caption{ 
    \textbf{Overview of the framework. Through logic-informed pretraining, \HHRL{} embeds logic priors directly into neural policies, thereby addressing the deep policy misalignment issue.} 
    \HHRL{} provides a two-stage training paradigm. In the first stage, the deep policy is jointly trained with the logic manager and the gating module (referred to as deep policy pretraining). In the second stage, the deep policy is further trained through direct interaction with the environment (referred to as deep policy post-training). 
    See Sec.~\ref{HHRL} for details.
    }
    \label{fig:structure}
\end{figure*}


\section{Background}
\label{background}
A key component of \HHRL{} is \emph{differentiable logic reasoning}; 
for a detailed review of first-order logic, we refer the reader to App.~\ref{FOL}.
Akin to \cite{Shindo21aaai, shindo2023alpha}, by defining the initial and $t$-th step valuation of ground atoms as $\mathbf{v}^{(0)}$ and $\mathbf{v}^{(t)}$, we make the logic manager and logic gating module differentiable in three steps: \textbf{(Step 1)} We encode each reasoning rule \( C_i \in \mathcal{C} \) as a tensor \( \mathbf{I}_i \in \mathbb{N}^{G \times S \times L} \), where \( S \) is the maximum number of possible substitutions for variables, \( L \) is the maximum number of body atoms and \( G \) is the number of grounded atoms. Specifically, the tensor \( \mathbf{I}_i \) stores at position $[j, k, l]$ the index (0 to $G-1$) of the grounded atom that serves as the $l$-th body atom when rule $C_i$ derives grounded head $j$ using substitution $k$. \textbf{(Step 2)} To be able to learn which rules are most relevant during forward reasoning, a weight matrix $\mathbf{W}$ consisting of $M$ learnable weight vectors, $[\mathbf{w}_1, \dots, \mathbf{w}_M]$, is introduced. Each vector $\mathbf{w}_m \in \mathbb{R}^C$ contains raw weights for the $C$ rules. To convert these raw weights into normalized probabilities for soft rule selection, a \textit{softmax} function is applied independently to each vector $\mathbf{w}_m$, yielding $\mathbf{w}^*_m$. \textbf{(Step 3)} At each step \( t \), we compute the valuation of body atoms using the \textit{gather} operation over the valuation vector \( \mathbf{v}^{(t)} \), looping over the body atoms for each grounded rule. These valuations are combined using a soft logical AND (\emph{gather} function) followed by a soft logical OR across substitutions:

\begin{equation}
        b_{i,j,k}^{(t)} = \prod\nolimits_{1 \leq l \leq L} \mathbf{gather}(\mathbf{v}^{(t)}, \mathbf{I}_i)[j,k,l], \
        c_{i,j}^{(t)} = \mathit{softor}^\gamma (b_{i,j,1}^{(t)}, \dots, b_{i,j,S}^{(t)}).
    \end{equation}

Here, \( i \) indexes the rule, \( j \) the grounded head atom, and \( k \) the substitution applied to existentially quantified variables. The resulting body evaluations \( c_{i,j}^{(t)} \) are weighted by their assigned rule weights \( w^*_{m,i} \), and then aggregated across rules and rule sets:
\begin{equation}
    h_{j,m}^{(t)} = \sum\nolimits_{1 \leq i \leq C} w^*_{m,i} \cdot c_{i,j}^{(t)}, \ \ 
    r_{j}^{(t)} = \mathit{softor}^\gamma ( h_{j,1}^{(t)}, \dots, h_{j,M}^{(t)} ), \ \
    v^{(t+1)}_j = \mathit{softor}^\gamma (r^{(t)}_j, v^{(t)}_j).
\end{equation}

We provide full details of this differentiable reasoning procedure in App.~\ref{diff}.

\section{Hybrid Hierarchical Reinforcement Learning}
\label{HHRL}
We now present the \HHRL{} framework.
As shown in Fig.~\ref{fig:structure}, \HHRL{} consists of four components:
(i) a differentiable symbolic logic manager, 
(ii) a set of pretrained option workers,
(iii) a neural RL policy, and 
(iv) an \emph{MoE} gating module.
\HHRL{} provides a two-stage training paradigm. In the first stage of training (pretraining), the neural RL policy is jointly trained with the logic manager and the gating module. In the second stage of training (posttraining), the neural RL policy is further trained through standard interactions with the environment. We now introduce the individual modules and explain how they are jointly trained. Specifically, we use PPO as an example of a neural RL policy.

We consider a Markov decision process (MDP) with state space $\mathcal{S}$, action space $\mathcal{A}$, transition kernel $p$, and reward function $\mathcal{R}$. In our environments, the agent observes both a low-level visual state $x_t \in \mathbb{R}^{4 \times 84 \times 84}$ (a stack of frames) and a high-level symbolic state $z_t$, for example, an object-centric representation. Together, they form the full state
$s_t = (x_t, z_t) \in \mathcal{S}\;.$

\textbf{Logic manager and pretrained option workers.}
The logic manager is a differentiable logic program, parameterized by $\theta$, that maps the symbolic state $z_t$ to a distribution over a finite set of option workers $\mathcal{O}:
    \pi_{\mathrm{logic}}(o_t \mid z_t; \theta)
    \quad \text{for } o_t \in \mathcal{O}.$ 
Each option worker $o \in \mathcal{O}$ corresponds to a low-level policy $\pi_{\omega_o}(a_t \mid x_t)$ that is trained separately on a subtask, such as ``grab the hammer'', ``use hammer'', or ``climb up''. The workers are not restricted to neural policies, but could also be logic-based. For details on how we pretrain the options, please refer to App.~\ref{app.options_pretrain}. These workers are kept fixed during agent training. Conditioned on $z_t$, the logic manager defines a hierarchical distribution over actions by marginalizing over options: $\pi_{\mathrm{L}}(a_t \mid x_t, z_t)=
    \sum\nolimits_{o \in \mathcal{O}} 
    \pi_{\mathrm{logic}}(o \mid z_t; \theta)\,
    \pi_{\omega_o}(a_t \mid x_t).
    \label{eq:logic-induced-policy}
$

\textbf{Neural RL policy and Gating Module (MoE).}
In parallel to the logic manager, the neural RL policy $\pi_{\mathrm{N}}(a_t \mid x_t; \phi)$ is jointly trained with the MoE Module on the raw visual input $x_t$. It uses the standard actor-critic architecture for Atari-style domains: a convolutional backbone followed by a linear policy head and a linear value head. The neural controller directly parameterizes a distribution over primitive actions: $a_t \sim \pi_{\mathrm{N}}(\cdot \mid x_t; \phi)\;.$
We introduce a gating module $b_\psi$, as a \emph{Mixture-of-Experts (MoE)} gate, to combine the output of the logic manager with the neural policy. This module $b_\psi$ outputs a distribution over \textit{logic} and \textit{neural} control. Depending on the configuration, $b_\psi$ can be:
(i) a differentiable logic program operating on $z_t$ (logic-based gating), or
(ii) a convolutional network operating on $x_t$ (neural-based gating).
We define: $
    \beta_t = (\beta_t^{\mathrm{L}}, \beta_t^{\mathrm{N}})
    = b_\psi(b_t),
$
with $\beta_t^{\mathrm{L}}, \beta_t^{\mathrm{N}} \ge 0$ and $\beta_t^{\mathrm{L}} + \beta_t^{\mathrm{N}} = 1$, where $b_t = z_t$ for logic gating and $b_t = x_t$ for neural gating. In practice, $b_\psi$ produces unnormalized logits, which are passed through a softmax to obtain $\beta_t$. Given the logic induced policy $\pi_{\mathrm{L}}(\cdot \mid x_t, z_t)$ in Eq.~\eqref{eq:logic-induced-policy} and the neural policy $\pi_{\mathrm{N}}(\cdot \mid x_t)$, the final policy is a convex combination, i.e., $\pi_{\mathrm{H}}(a_t \mid x_t, z_t) = 
 \beta_t^{\mathrm{L}}\,\pi_{\mathrm{L}}(a_t \mid x_t, z_t)
    + \beta_t^{\mathrm{N}}\,\pi_{\mathrm{N}}(a_t \mid x_t).
    \label{eq:hybrid-policy}
$



\textbf{Value function.}
The hybrid agent also maintains a value function that combines a logic critic and a neural critic. The logic critic $V_{\mathrm{L}}(z_t; \theta_V)$ is implemented as an MLP on symbolic features, while the neural critic $V_{\mathrm{N}}(x_t; \phi_V)$ shares the convolutional backbone with the PPO policy network. Consistent with the policy mixture we define
$
    V_{\mathrm{H}}(s_t)
    = \beta_t^{\mathrm{L}}\,V_{\mathrm{L}}(z_t; \theta_V)
    + \beta_t^{\mathrm{N}}\,V_{\mathrm{N}}(x_t; \phi_V).
    \label{eq:hybrid-value}
$

\textbf{Training objective.}
We optimize the parameters of the neural PPO, the logic critic, and the gating module jointly. 
Let $\theta$ denote all trainable parameters of the hybrid agent (including $\phi$, $\theta_V$, $\phi_V$, and, when trainable, the logic manager and gating parameters). For a batch of trajectories, we compute standard generalized advantage estimates $A_t$ and returns $R_t$. The surrogate for the hybrid policy 
is $
    L_{\mathrm{clip}}(\theta)
    =
    \mathbb{E}_t \Big[
        \min\big(
            r_t(\theta) A_t,\,
            \operatorname{clip}(r_t(\theta), 1 - \epsilon, 1 + \epsilon) A_t
        \big)
    \Big],\ \text{where} \ 
    r_t(\theta)
    =
    \tfrac{
        \pi_{\mathrm{H}}(a_t \mid x_t, z_t; \theta)
    }{
        \pi_{\mathrm{H}}^{\mathrm{old}}(a_t \mid x_t, z_t)
    }.
$
The hybrid value function 
is optimized with a squared error loss: 
$
    L_{\mathrm{V}}(\theta)
    = \mathbb{E}_t \bigl[ \bigl( V_{\mathrm{H}}(s_t; \theta) - R_t \bigr)^2 \bigr].
$

To encourage exploration, we add two entropy regularizers: 
(i) the entropy of the action distribution $H\bigl(\pi_{\mathrm{H}}(\cdot \mid x_t, z_t)\bigr)$, and 
(ii) the entropy of the gating distribution $H(\beta_t)$. 
The full loss
\begin{equation}
\begin{aligned}
        \mathcal{L}(\theta)
    = - L_{\mathrm{clip}}(\theta)
      +c_V\,L_{\mathrm{V}}(\theta)
      -c_{\mathrm{ent}}\,\mathbb{E}_t\big[H\big(\pi_{\mathrm{H}}(\cdot \mid x_t, z_t)\big)\big]
      - c_{\mathrm{gate}}\,\mathbb{E}_t\big[H(\beta_t)\big],
\end{aligned}
\end{equation}
is minimized during training.
The coefficients $c_V$, $c_{\mathrm{ent}}$, and $c_{\mathrm{gate}}$ correspond to the value loss, action entropy, and gating entropy, respectively. 

This completes the pretraining stage. So far, we have two policies: the full \HHRL{} framework policy and the neural component part of \HHRL{}, the \hhrlp{} policy. We then posttrain \hhrlp{} via standard on-policy interaction with the environment, which results in the policy \hhrlpp.


\section{Experimental Evaluation}
\label{experiments}

With \HHRL{} at hand, our intention is now to investigate five research questions: \textbf{(RQ1)} How does \HHRL{} perform compared with baselines? \textbf{(RQ2)} Can \HHRL{} pretraining be utilized to boost other deep RL methods? \textbf{(RQ3)} Can \HHRL{} pretraining successfully address policy misalignment? \textbf{(RQ4)} How do the different components affect the performance of \HHRL? \textbf{(RQ5)} Does \HHRL{} scale to continuous action spaces? 

\textbf{Experimental setup.}
We evaluate all methods on the Atari Learning Environment (ALE)~\citep{bellemare2013arcade} with discrete action spaces, including \textit{Seaquest} and two more challenging games \textit{Kangaroo} and \textit{DonkeyKong}, which present long-horizon dependencies and reward traps. Additionally, we assess our method and baseline approaches in the Continuous Atari Learning Environment (CALE)~\citep{farebrother2024cale}, focusing on the two more challenging games, \textit{Kangaroo} and \textit{DonkeyKong}, with continuous action space. Observations are preprocessed using the standard Atari pipeline: frames are converted to grayscale, resized to $84 \times 84$, and stacked over the last 4 time steps, yielding a visual input 
$x_t$. For the logic manager, the symbolic state $z_t$ is obtained through OCAtari~\citep{ocatari} and used only during pretraining. Note that, in \textit{Kangaroo} and \textit{DonkeyKong}, the game does not advance beyond the first level; it restarts at the first level after ending.

\textbf{Baseline Methods.}
We compare \HHRL{} against neural, hierarchical, and neuro-symbolic baselines, including: PPO~\citep{schulman2017proximal}, DQN~\citep{mnih2015human}, NUDGE~\citep{delfosse2023interpretable}, BlendRL~\citep{shindoblendrl}, Option-critic~\citep{bacon2017option}, C51~\citep{c51}, and hDQN~\citep{kulkarni2016hierarchical}. All variants and baseline models are trained and evaluated under identical environment settings to ensure fair comparison. Since no public open-source implementation of hDQN was available, we developed a custom version, with architecture details in App.~\ref{app:hdqn-arch}. Due to instabilities encountered during the evaluation phase, we reported the maximum performance achieved during training as a representative metric. Additionally, while we attempted to include Option-critic, existing public implementations do not natively support the \textit{DonkeyKong} environment. Our efforts to adapt the framework were unsuccessful; therefore, we exclude Option-critic for \textit{DonkeyKong}. All methods were trained on an RTX 2090 GPU with 12 GB.

\textbf{Logic manager and pretrained options.} We provide the logic manager, gating rules in App.~\ref{app:logicrules}. We pretrain the logic options using HackAtari. Details on how to train individual options are provided in App.~\ref{app.options_pretrain}. Note that the logic manager rules and the individual options may not be optimal.  

\textbf{Two-Stage Training in \HHRL.}
\textbf{(1) Pretraining:} We jointly train the gating module and neural RL policy while the logic manager selects from a fixed set of pretrained options via argmax, yielding \textbf{\HHRL{}} (full hybrid policy) and \textbf{\hhrlp} (neural policy part).
\textbf{(2) Posttraining:} We further train \hhrlp{} via standard environment interaction to obtain \textbf{\hhrlpp}. Unless stated otherwise, \HHRL{} uses PPO as the default neural policy.

\begin{figure*}[t]
    \centering
    \includegraphics[width=\textwidth]{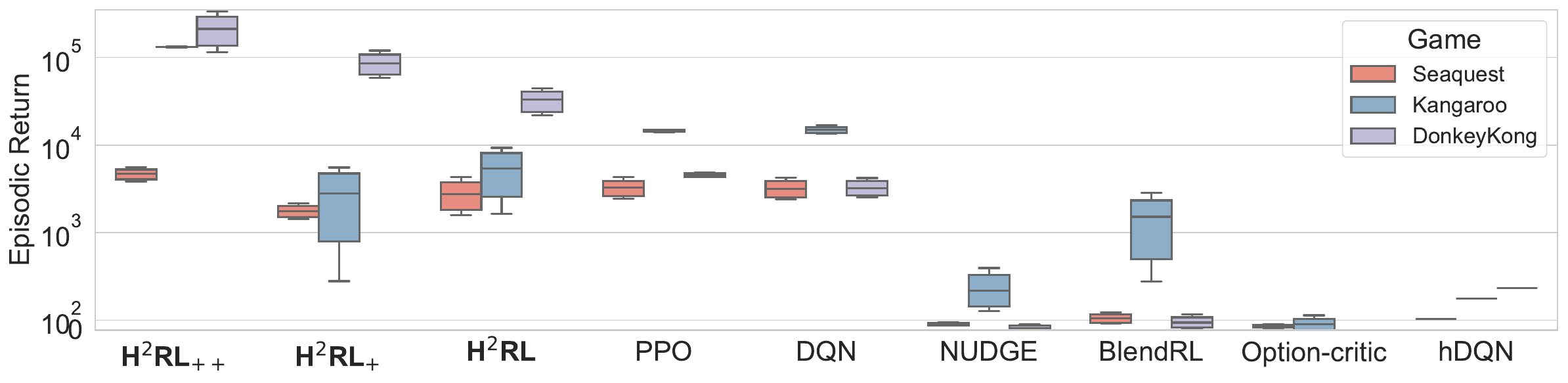}
    \caption{
    \textbf{Leveraging logic-informed pretraining, \HHRL{} with its variants (bolded), outperforms baselines on challenging ALE tasks (\textit{Seaquest}, \textit{Kangaroo}, and \textit{DonkeyKong}) with long-horizon dependencies and reward traps.} Although DQN and PPO achieve high returns in \textit{Kangaroo}, their learned policies remain misaligned; see Sec.~\ref {experiments} (RQ1 and RQ3) for details. Episodic returns are averaged over 12 environments (with 200 runs per environment). Results are presented on a log scale to normalize for the disparate reward magnitudes across games.
    } \label{fig:seaquestresult}
\end{figure*}
\textbf{RQ1:} \HHRL{} with its variants demonstrate a significant leap in performance and outperform all compared baselines. We evaluate the performance of three variants of \HHRL{} against SOTA deep RL, hierarchical RL, and neuro-symbolic RL baselines across three challenging Atari environments. The results, summarized in Fig.~\ref{fig:seaquestresult} (and in App.~Tab.~\ref{tab:rq1table}). 
Most notably, \hhrlpp{} achieves scores that are orders of magnitude higher than the baselines, reaching 131,842\spm{1,221} in \textit{Kangaroo} and 216,793\spm{125,655} in \textit{DonkeyKong}. These environments are notorious for misleading agents with early, dense rewards. 
The fact that \HHRL{} maintains its trajectory toward high-level objectives, where other agents plateau, suggests that our logic-informed pretraining effectively boosts the agents' performance. Furthermore, the performance jump from the pretraining policy \hhrlp{} to the post-training policy \hhrlpp{} validates our two-stage approach, showing that the symbolic scaffolding provides a foundation upon which the neural policy can eventually achieve peak performance that neither purely symbolic nor deep RL methods can reach alone.
\begin{table*}[t]
\small
\centering
\caption{ \textbf{\HHRL{} serves as a universal pretraining substrate for on-policy and off-policy deep RL methods.} Shown are the means of the episodic return (averaged over 12 environments, 200 runs per environment) with the std of the pretrained and base policy. Although DQN and PPO achieve high returns in \textit{Kangaroo}, their learned policies remain misaligned; Details see Sec.~\ref {experiments} (RQ3).}
\begin{tabular}{lcccccccc}
\toprule
            & H$^2$PPO$_{+}$              & PPO           & H$^2$DQN$_{+}$            & DQN                      & H$^2$C51$_{+}$                 & C51  \\
\midrule
Seaquest    & 1802\spm{400}     & \textbf{3247\spm{881}}  & 1966\spm{903}              & \textbf{3259\spm{1001}}  & 2774\spm{1367}              & \textbf{4381}\spm{1988}  \\
Kangaroo    & 2754\spm{2626}   & \textbf{14592\spm{491}} & \textbf{114665\spm{16932}} & 14822\spm{1810}          & 8193\spm{5928}              & \textbf{13854}\spm{1021}  \\
DonkeyKong  & \textbf{87780\spm{32786}} & 4536\spm{296} & \textbf{6887\spm{1765}}    & 3205\spm{988}            & \textbf{183645\spm{107378}} & 3393\spm{721}  \\
\bottomrule
\end{tabular}
\label{tab:rq2table}
\end{table*}


\begin{table*}[t]
\small
\centering
\caption{\textbf{\HHRL{} mitigates policy misalignment via logic-informed pretraining for on-policy and off-policy deep RL methods.} The table reports success rates for reaching different floors in \textit{Kangaroo} for agents pre-trained using \HHRL{}, its base variant, and the logic manager (hReason). Success rates averaged over 5 seeds, with 6 runs per seed.}
\begin{tabular}{lcccccccc}
\toprule
            & PPO                                & DQN                                  & C51                                  & hReason                                  & H$^2$PPO                                 & H$^2$DQN$_{+}$                                 & H$^2$C51$_{+}$                                 \\
\midrule
Floor 2   & \cellcolor{OliveGreen!0}0\%\spm{0\%} & \cellcolor{OliveGreen!0}0\%\spm{0\%} & \cellcolor{OliveGreen!0}0\%\spm{0\%} & \cellcolor{OliveGreen!60}100\%\spm{0\%} & \cellcolor{OliveGreen!60}100\%\spm{0\%} & \cellcolor{OliveGreen!60}100\%\spm{0\%} & \cellcolor{OliveGreen!60}100\%\spm{0\%} \\
Floor 3   & \cellcolor{OliveGreen!0}0\%\spm{0\%} & \cellcolor{OliveGreen!0}0\%\spm{0\%} & \cellcolor{OliveGreen!0}0\%\spm{0\%} & \cellcolor{OliveGreen!18}30\%\spm{20\%}  & \cellcolor{OliveGreen!36}60\%\spm{10\%}  & \cellcolor{OliveGreen!60}100\%\spm{0\%} & \cellcolor{OliveGreen!60}100\%\spm{0\%} \\
Floor 4   & \cellcolor{OliveGreen!0}0\%\spm{0\%} & \cellcolor{OliveGreen!0}0\%\spm{0\%} & \cellcolor{OliveGreen!0}0\%\spm{0\%} & \cellcolor{OliveGreen!18}30\%\spm{10\%}  & \cellcolor{OliveGreen!30}50\%\spm{10\%}  & \cellcolor{OliveGreen!60}100\%\spm{0\%} & \cellcolor{OliveGreen!60}100\%\spm{0\%} \\
\bottomrule
\end{tabular}
\label{tab:generalpretraining}
\end{table*}

\textbf{RQ2:} \HHRL{} can serve as a universal pretraining substrate for both on-policy and off-policy methods. For off-policy pretraining, we use the logic manager to collect a replay buffer and then train the agent on it. Tab~\ref{tab:rq2table} compares the average episodic returns of pretrained methods with their base variants in \textit{Seaquest}, \textit{Kangaroo}, and \textit{DonkeyKong}. Overall, \HHRL{} substantially improves episodic returns in games with long-horizon dependencies such as \textit{DonkeyKong}. In contrast, for simpler games such as \textit{Seaquest}, we do not observe a clear pretraining gain, which may stem from non-optimal logic-manager design or suboptimal option definitions. We note that, despite achieving high returns in \textit{Kangaroo}, DQN and PPO still learn misaligned policies as shown in RQ3. 

\textbf{RQ3:} \HHRL{} pretraining can mitigate policy misalignment. We report success rates for reaching different floors in \textit{Kangaroo} for PPO, DQN, and C51, as well as their \HHRL{}-pretrained variants. As shown in Tab.~\ref{tab:generalpretraining}, rather than getting stuck in the corner of the base variants (Fig.~\ref{fig:ppofail}), all \HHRL{}-pretrained agents (across on-policy and off-policy) successfully overcome the policy misalignment issue and consistently climb upward, outperforming their base variants. 

\textbf{RQ4:} \label{rq3}The integration of logic guidance and neural flexibility in \HHRL{} is crucial for its success. We conduct an ablation study to compare H$^2$PPO (the default version of \HHRL) with the neural-only part (PPO), the pure logic manager (hReason), a hierarchical neural manager (hPPO), and an augmented PPO provided with both pixel and symbolic data (exPPO). We provide exPPO's architecture in App.\ref{app:exppoarch}. As shown in Tab.~\ref{tab:rq3}, while exPPO and hPPO achieve high scores in Kangaroo (14,247\spm{1,085} and 10,601\spm{914}, respectively), they got trapped in the short-term reward by attacking the enemies in a corner (indicated by the success rate in Tab.~\ref{tab:rq3}). Instead, H$^2$PPO successfully avoids reward hacking and proceeds along the intended game path without getting stuck in the corner to collect points by attacking enemies. This logic-guided game progression can significantly improve the return, for example, in \textit{DonkeyKong}, where H$^2$PPO reaches 33,657\spm{14,578}, outperforming the next-closest PPO by nearly an order of magnitude. In this environment, the pure logic manager (hReason) and the hierarchical neural manager (hPPO) both fail significantly, suggesting that neither symbolic nor hierarchical structure is sufficient in isolation. Furthermore, the fact that exPPO, which has access to the same symbolic information as H$^2$PPO but lacks our logic-informed pretraining, fails to match H$^2$PPO's performance in \textit{Seaquest} and \textit{DonkeyKong} confirms that it is insufficient to simply provide symbolic state information. Instead, the results demonstrate that \HHRL's two-stage framework, which uses logic to steer learning toward goal-directed behavior, is the critical factor in overcoming long-horizon decision-making challenges that individual components alone cannot solve.
\begin{table*}[t]
\small
\centering
\caption{
\textbf{The synergy between the logic and neural modules is crucial to \HHRL’s improved performance.} As an ablation study, we compare H$^2$PPO (the default \HHRL{}) with PPO (neural-only), hPPO (neural manager), hReason (logic manager), and exPPO (PPO with extended inputs) on \textit{Seaquest}, \textit{Kangaroo}, and \textit{DonkeyKong}. The table reports the average episodic returns and success rates for reaching the third floor in \textit{Kangaroo}. For details, see Sec.~\ref {rq3} (RQ3).}
\begin{tabular}{lcccccc}  
\toprule 
              & H$^2$PPO                                     
              & PPO                                   & hPPO                                   & hReason                                & exPPO                                   \\
\midrule
Seaquest      & \cellcolor{OliveGreen!47}2812\spm{1477}  
              & \cellcolor{OliveGreen!60}3247\spm{881}
              & \cellcolor{OliveGreen!19}1906\spm{628}
              & \cellcolor{OliveGreen!0}1281\spm{963} 
              & \cellcolor{OliveGreen!17}1848\spm{34}
              \\
Kanga./Succ.     & \cellcolor{OliveGreen!15}5351\spm{4132}/ 0.6\spm{0.1}  
                 & \cellcolor{OliveGreen!60}14592\spm{491}/ 0 
                 & \cellcolor{OliveGreen!41}10601\spm{914}/ 0
                 & \cellcolor{OliveGreen!0}2238\spm{2140}/ 0.3\spm{0.2}
                 & \cellcolor{OliveGreen!58}14247\spm{1085}/ 0
                 \\
DonkeyKong    & \cellcolor{OliveGreen!60}33657\spm{14578} 
              & \cellcolor{OliveGreen!7}4536\spm{296}
              & \cellcolor{OliveGreen!0}418\spm{139}
              & \cellcolor{OliveGreen!1}905\spm{1335}
              & \cellcolor{OliveGreen!7}4268\spm{249}
              \\
\bottomrule
\end{tabular}
\label{tab:rq3}
\end{table*}
\begin{wrapfigure}[20]{t}{0.4\textwidth}
    \centering
    \includegraphics[width=0.4\textwidth]{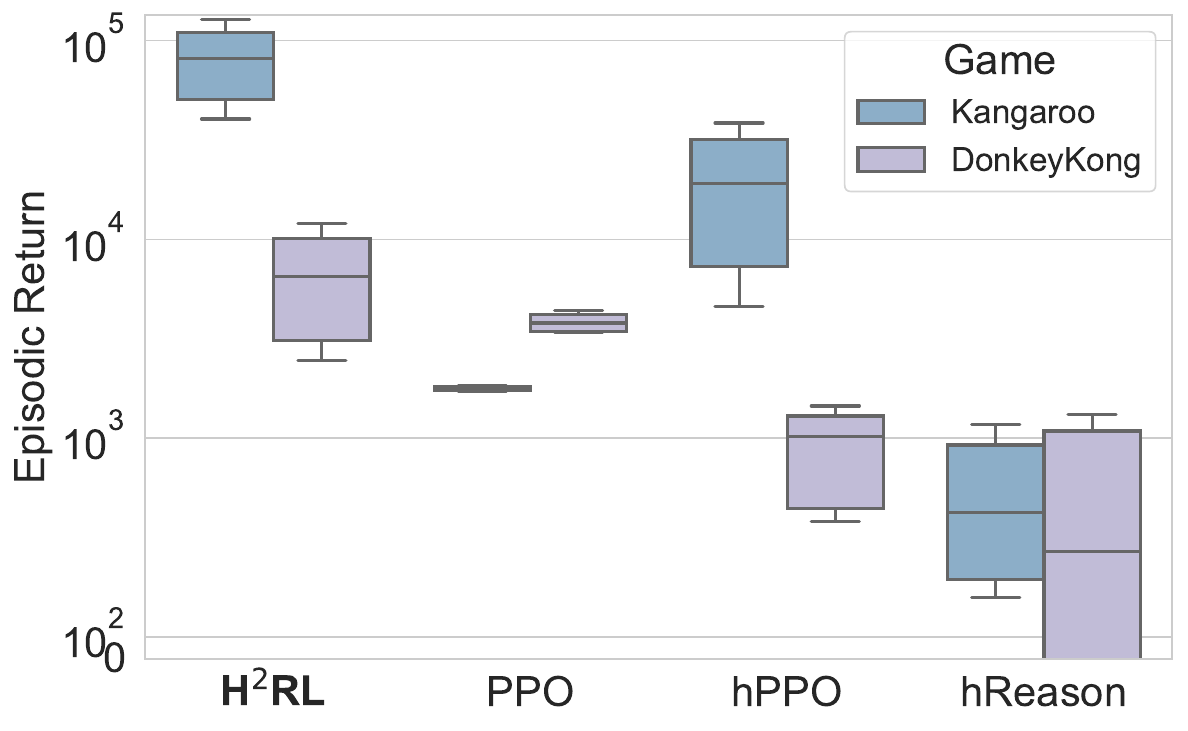}
    \caption{\textbf{\HHRL{} effectively leverages logic reasoning in continuous action spaces and improves deep agents' performance.} 
    We compare \HHRL{} with methods applicable to continuous action space 
    on the \textit{Kangaroo} and \textit{DonkeyKong} tasks in CALE, where \HHRL{} consistently outperforms these baselines. Details see Sec.~\ref{experiments}: RQ4.}
    \label{fig:cale}
\end{wrapfigure}

\textbf{RQ5:} H$^2$RL effectively leverages logic reasoning in continuous action spaces and improves deep agents' performance. 
We evaluate \HHRL{} within the Continuous Atari Learning Environment (CALE) to determine whether the integration of symbolic structures benefits agents in non-discrete domains. The results, illustrated in Fig.~\ref{fig:cale} (and in Tab.~\ref{tab:rq5table}), demonstrate that \HHRL{} pretraining significantly improves upon the baseline PPO agent across diverse tasks (we compare against baselines that are able to apply to continuous-action settings, namely PPO, hPPO, and hReason). Specifically, in \textit{Kangaroo (cont.)} environment, \HHRL{} achieves a mean score of 84,665\spm{49,767}, substantially outperforming both the PPO baseline (1,785\spm{72}) and the hierarchical variant, hPPO (19,854\spm{18,586}). This performance gap confirms that logic-informed pretraining provides a decisive advantage in continuous action space, where the structured guidance provided by logic is not confined to discrete settings, rather, it is equally potent for continuous action space.

\textbf{Discussion:} The experimental results across RQ1–RQ5 provide a comprehensive validation of \HHRL's effectiveness to mitigate policy misalignment and as a universal pretraining framework. By outperforming neural and neuro-symbolic baselines by orders of magnitude in both discrete and continuous action spaces, \HHRL{} demonstrates that symbolic scaffolding is a powerful tool for maintaining a trajectory toward high-level objectives. The ablation study further clarifies that this success is not a byproduct of simple data augmentation. Neither providing extended symbolic information to a neural agent 
nor using a purely hierarchical neural manager could replicate \HHRL’s performance. This confirms that the logic-informed pretraining is the critical mechanism for overcoming long-horizon challenges. Finally, our extensions to off-policy methods and to continuous action spaces underscore the framework's algorithmic flexibility.  
Collectively, these results position \HHRL{} not merely as a novel RL algorithm, but as a versatile architectural paradigm that successfully bridges high-level reasoning and low-level control.

\section{Related Work}
\label{related work}
\textbf{Neuro-symbolic RL.} Neuro-symbolic
RL seeks to combine neural function approximation with symbolic reasoning for improved generalization and interpretability. 
Related efforts include NLRL~\citep{jiang2019neural}, Galois~\citep{cao2022galois}, ESPL~\citep{guo2023efficient}, and BlendRL~\citep{shindoblendrl}, however, they focus on synthesizing symbolic or logical policies that map directly to raw actions.
Instead of raw actions, hierarchical RL~\citep{sutton1999between, dietterich2000hierarchical} allows to decompose tasks into subtasks or options~\citep{sutton1999between,bacon2017option, vezhnevets2017feudal}. NEXUS~\citep{nexus} and Dylan~\citep{dylan} employ (neuro-)symbolic meta-policies to guide such neural options.
In contrast, \HHRL{} is a pretraining framework with a differentiable logic reasoner, allowing deep RL agents to inherit logic priors through pretraining without the computational burden of symbolic reasoning during test-time inference.

\textbf{Imitation learning.} Imitation learning (IL), such as GAIL~\citep{gail} and learning from demonstration, has a long history in machine learning~\citep{bain1995framework}. While many prior works leverage expert demonstrations~\citep{cheng2020, ilhan2021student}, most existing methods assume access to a high-quality expert or teacher~\citep{derstroff2024peer}. In contrast, the Logic Manager in our framework selects among pretrained options and is explicitly non-expert (see~Tab.\ref{tab:rq3}, hReason). Related work has also explored learning from sparse or budgeted teacher feedback~\citep{ilhan2021student} and learning whom to trust~\citep{nunes2003}. Several works combine imitation and RL by pretraining or regularizing policies with behavioral cloning (BC). For example, \citet{rajeswaran2017learning} pretrains a policy via BC before RL, while \citet{goecks2020behavioralcloning} and \citet{huang2023guided} incorporate an auxiliary BC loss and sample expert trajectories from a dedicated replay buffer. In contrast, \HHRL{} does not rely on recorded human demonstrations or expert policies. Instead, it injects heuristic guidance into the neural policy through logic-informed pretraining.

\textbf{Shortcut mitigation.} 
To address the fragility of reward-driven agents, current mitigation strategies typically frame robustness as one of their goals~\citep{PintoDSG17,farebrother18generalization,Delfosse25hackatari}. Existing strategies are either extrinsic methods, such as domain randomization~\citep{TobinFRSZA17} and image augmentations~\citep{yarats2021image}, which increase data diversity to force invariance, or algorithmic solutions utilizing auxiliary objectives for invariant feature embeddings~\citep{zhang2021learning, bertoin2022look}.
%
We diverge from these approaches by using symbolic \textit{scaffolding} (cf. \citet{bransford2000people}) to create a new training setup. Unlike classical neurosymbolic approaches that suffer from a ``latency bottleneck'' due to continuous symbolic reasoning~\citep{Bhuyan24nesysurvey}, \HHRL{} directly embeds a structural inductive bias into the neural policy 
while retaining the efficiency of a neural approach. 

\textbf{Exploration.} Effective exploration in deep RL typically relies on either stochastic noise or intrinsic motivation. Foundational methods like $\epsilon$-greedy~\citep{mnih2015human} and entropy maximization~\citep{haarnoja2018soft} introduce random perturbations to ensure coverage, but they often result in unstructured and inefficient behavior in complex environments. To drive more directed discovery, sophisticated approaches augment the reward signal with ``novelty bonuses''. This includes count-based methods~\citep{bellemare2016unifying}, curiosity-driven prediction errors~\citep{pathak2017curiosity}, and Random Network Distillation~\citep{burda2019exploration}.
In contrast, \HHRL{} approaches exploration as a structural side effect rather than an explicit objective.
By employing a hierarchical structure governed by a logical meta-policy, 
the resulting policy benefits from the structured guidance during the logic-informed pretraining phase, 
without requiring complex intrinsic reward signals.

\section{Conclusions}
\label{conclusions}
We proposed \HHRL, a novel hierarchical neuro-symbolic reinforcement learning framework that leverages logic-informed pretraining to steer deep RL policies away from misalignment. Empirically, we demonstrate that \HHRL{} significantly enhances deep RL agents' performance on challenging, long-horizon tasks while requiring no logical reasoning at inference time. Furthermore, \HHRL{} can serve as a universal pretraining substrate for both on-policy and off-policy methods, providing a pathway for integrating logical reasoning into reinforcement learning in continuous action spaces. In future work, we aim to integrate \HHRL{} into real-world robotic systems, where structured priors and safety-aware reasoning are critical. Additionally, we plan to scale the framework to more complex decision-making environments involving high-dimensional observations and adaptive, multi-level reasoning mechanisms.

\section{Acknowledgements}
\label{ackowledgements}
This work was funded by the Deutsche Forschungsgemeinschaft (DFG, German Research Foundation) under Germany´s Excellence Strategy – EXC-3066, ``The Adaptive Mind'', and EXC-3057, ``Reasonable AI''. It was also funded by the German Federal Ministry of Education and Research, the Hessian Ministry of Higher Education, Research, Science and the Arts (HMWK) within their joint support of the National Research Center for Applied Cybersecurity ATHENE, via the ``SenPai: XReLeaS'' project. The authors gratefully acknowledge the computing time provided to them on the high-performance computer Lichtenberg at the NHR Center NHR4CES@TUDa. This is funded by the German Federal Ministry of Education and Research (BMBF) and the Hessian Ministry of Science and Research, Art and Culture (HMWK). This project has been supported by a hardware donation by NVIDIA through the Academic Grant Program.
\bibliography{main}

@String{NeurIPS = {Advances in Neural Information Processing Systems ({NeurIPS})}}

@String{ICLR = {Proceedings of the International Conference on Learning Representations ({ICLR})}}

@String{ICML = {Proceedings of the International Conference on Machine Learning ({ICML})}}

@String{IJCAI={Proceedings of the International Joint Conference on Artificial Intelligence ({IJCAI})}}

@String{AAAI = {Proceedings of the {AAAI} Conference on Artificial Intelligence (AAAI)}}

@String{IROS = {IEEE/RSJ International Conference on Intelligent Robots and Systems (IROS)}}

@String{AAMAS = {Proceedings of the International Conference on Autonomous Agents and Multiagent Systems (AAMAS)}}

@String{RLJ = {Reinforcement Learning Journal ({RLJ})}}

@String{NMI = {Nature Machine Intelligence}}

@String{JAIR = {Journal of Artificial Intelligence Research ({JAIR})}}

@String{JMLR = {Journal of Machine Learning Research ({JMLR})}}

@String{MLJ = {Machine Learning (MLJ)}}

@article{mnih2015human,
  title={Human-level control through deep reinforcement learning},
  author={Mnih, Volodymyr and Kavukcuoglu, Koray and Silver, David and Rusu, Andrei A and Veness, Joel and Bellemare, Marc G and Graves, Alex and Riedmiller, Martin and Fidjeland, Andreas K and Ostrovski, Georg and others},
  journal={Nature},
  year={2015},
}

@article{schulman2017proximal,
  title={Proximal policy optimization algorithms},
  author={Schulman, John and Wolski, Filip and Dhariwal, Prafulla and Radford, Alec and Klimov, Oleg},
  journal={arXiv preprint arXiv:1707.06347},
  year={2017}
}

@inproceedings{pathak2017curiosity,
  title={Curiosity-driven exploration by self-supervised prediction},
  author={Pathak, Deepak and Agrawal, Pulkit and Efros, Alexei A and Darrell, Trevor},
  booktitle=ICML,
  year={2017},
}

@inproceedings{Shindo21aaai,
  author    = {Hikaru Shindo and
               Masaaki Nishino and
               Akihiro Yamamoto},
  title     = {Differentiable Inductive Logic Programming for Structured Examples},
  booktitle = AAAI,
  year      = {2021},
}

@book{Russel09,
author = {Russell, Stuart and Norvig, Peter},
title = {Artificial Intelligence: A Modern Approach},
year = {1995},
publisher = {Prentice Hall Press},
}

@inproceedings{delfosse2023interpretable,
  title={Interpretable and explainable logical policies via neurally guided symbolic abstraction},
  author={Delfosse, Quentin and Shindo, Hikaru and Dhami, Devendra and Kersting, Kristian},
  booktitle=NeuRIPS,
  year={2023}
}

@inproceedings{jiang2019neural,
  title={Neural logic reinforcement learning},
  author={Jiang, Zhengyao and Luo, Shan},
  booktitle=ICML,
  year={2019},
}

@article{shindo2023alpha,
  title={$\alpha$ ilp: thinking visual scenes as differentiable logic programs},
  author={Shindo, Hikaru and Pfanschilling, Viktor and Dhami, Devendra Singh and Kersting, Kristian},
  journal=MLJ,
  year={2023},
}

@inproceedings{cuturi2017soft,
  title={Soft-dtw: a differentiable loss function for time-series},
  author={Cuturi, Marco and Blondel, Mathieu},
  booktitle=ICML,
  year={2017}
}

@inproceedings{kulkarni2016hierarchical,
  title={Hierarchical deep reinforcement learning: Integrating temporal abstraction and intrinsic motivation},
  author={Kulkarni, Tejas D and Narasimhan, Karthik and Saeedi, Ardavan and Tenenbaum, Josh},
  booktitle=NeuRIPS,
  year={2016}
}

@article{ye2022differentiable,
  title={Neural Meta-Symbolic Reasoning and Learning},
  author={Ye, Zihan and Shindo, Hikaru and Dhami, Devendra Singh and Kersting, Kristian},
  journal={arXiv preprint arXiv:2211.11650},
  year={2022}
}

@inproceedings{cao2022galois,
  title={GALOIS: boosting deep reinforcement learning via generalizable logic synthesis},
  author={Cao, Yushi and Li, Zhiming and Yang, Tianpei and Zhang, Hao and Zheng, Yan and Li, Yi and Hao, Jianye and Liu, Yang},
  booktitle=NeuRIPS,
  year={2022}
}

@inproceedings{guo2023efficient,
  title={Efficient symbolic policy learning with differentiable symbolic expression},
  author={Guo, Jiaming and Zhang, Rui and Peng, Shaohui and Yi, Qi and Hu, Xing and Chen, Ruizhi and Du, Zidong and Zhang, Xishan and Li, Ling and Guo, Qi and Chen, Yunji},
  booktitle=NeuRIPS,
  year={2023}
}

@article{bellemare2013arcade,
  title={The arcade learning environment: An evaluation platform for general agents},
  author={Bellemare, Marc G and Naddaf, Yavar and Veness, Joel and Bowling, Michael},
  journal=JAIR,
  year={2013}
}

@inproceedings{shindoblendrl,
  title={BlendRL: A Framework for Merging Symbolic and Neural Policy Learning},
  author={Shindo, Hikaru and Delfosse, Quentin and Dhami, Devendra Singh and Kersting, Kristian},
  booktitle=ICLR,
  year= 2025
}

@article{
dylan,
title={Learning from Less: Guiding Deep Reinforcement Learning with Differentiable Symbolic Planning},
author={Zihan Ye and Oleg Arenz and Kristian Kersting},
journal={RLC 2025 Workshop on Programmatic Reinforcement Learning},
year={2025}
}

@inproceedings{goecks2020behavioralcloning,
author = {Goecks, Vinicius G. and Gremillion, Gregory M. and Lawhern, Vernon J. and Valasek, John and Waytowich, Nicholas R.},
title = {Integrating Behavior Cloning and Reinforcement Learning for Improved Performance in Dense and Sparse Reward Environments},
year = {2020},
booktitle=AAMAS,
}

@inproceedings{derstroff2024peer,
title={Peer Learning: Learning Complex Policies in Groups from Scratch via Action Recommendations},
booktitle=AAAI,
author={Derstroff, Cedric and Cerrato, Mattia and Brugger, Jannis and Peters, Jan and Kramer, Stefan},
year={2024},
}

@article{farebrother18generalization,
  author       = {Jesse Farebrother and
                  Marlos C. Machado and
                  Michael Bowling},
  title        = {Generalization and Regularization in {DQN}},
  year         = {2018},
  journal = {arXiv preprint arXiv:1810.00123}
}

@inproceedings{Hermann24shortcut,
  author       = {Katherine L. Hermann and
                  Hossein Mobahi and
                  Thomas Fel and
                  Michael Curtis Mozer},
  title        = {On the Foundations of Shortcut Learning},
  booktitle    = ICLR,
  year         = {2024}
}

@article{Delfosse25hackatari,
  author       = {Quentin Delfosse and
                  Jannis Bl{\"{u}}ml and
                  Fabian Tatai and
                  Th{\'{e}}o Vincent and
                  Bjarne Gregori and
                  Elisabeth Dillies and
                  Jan Peters and
                  Constantin A. Rothkopf and
                  Kristian Kersting},
  title        = {Deep Reinforcement Learning Agents are not even close to Human Intelligence},
  journal      = {arXiv preprint arXiv:2505.21731},
  year         = {2025}
}

@article{koch21objective,
  author       = {Jack Koch and
                  Lauro Langosco and
                  Jacob Pfau and
                  James Le and
                  Lee Sharkey},
  title        = {Objective Robustness in Deep Reinforcement Learning},
  journal      = {arXiv preprint arXiv:2105.14111},
  year         = {2021}
}

@article{Geirhos20Shortcuts,
  author       = {Robert Geirhos and
                  J{\"{o}}rn{-}Henrik Jacobsen and
                  Claudio Michaelis and
                  Richard S. Zemel and
                  Wieland Brendel and
                  Matthias Bethge and
                  Felix A. Wichmann},
  title        = {Shortcut learning in deep neural networks},
  journal      = NMI,
  volume       = {2},
  number       = {11},
  year         = {2020}
}

@inproceedings{Chan20reliable,
  author       = {Stephanie C. Y. Chan and
                  Samuel Fishman and
                  Anoop Korattikara and
                  John F. Canny and
                  Sergio Guadarrama},
  title        = {Measuring the Reliability of Reinforcement Learning Algorithms},
  booktitle    = ICLR,
  year         = {2020}
}

@inproceedings{Ilyas19adversarial,
  author       = {Andrew Ilyas and
                  Shibani Santurkar and
                  Dimitris Tsipras and
                  Logan Engstrom and
                  Brandon Tran and
                  Aleksander Madry},
  title        = {Adversarial Examples Are Not Bugs, They Are Features},
  booktitle    = NeuRIPS,
  year         = {2019}
}

@inproceedings{
skalse2022defining,
title={Defining and Characterizing Reward Gaming},
author={Joar Max Viktor Skalse and Nikolaus H. R. Howe and Dmitrii Krasheninnikov and David Krueger},
booktitle=NeuRIPS,
year={2022},
}

@article{Everitt21tampering,
  author       = {Tom Everitt and
                  Marcus Hutter and
                  Ramana Kumar and
                  Victoria Krakovna},
  title        = {Reward tampering problems and solutions in reinforcement learning:
                  a causal influence diagram perspective},
  journal      = {Synthese},
  volume       = {198},
  number       = {27},
  year         = {2021}
}

@book{Goodhart1984,
author="Goodhart, C. A. E.",
title="Monetary Theory and Practice: The UK Experience",
year="1984",
publisher="Macmillan Education UK",
isbn="978-1-349-17295-5",
}

@article{Grandien24interpretable,
  author       = {Nils Grandien and
                  Quentin Delfosse and
                  Kristian Kersting},
  title        = {Interpretable end-to-end Neurosymbolic Reinforcement Learning agents},
  journal      = {arXiv preprint arXiv:2410.14371},
  year         = {2024}
}

@inproceedings{Luo24insight,
  title={End-to-End Neuro-Symbolic Reinforcement Learning with Textual Explanations},
  author={Luo, Lirui and Zhang, Guoxi and Xu, Hongming and Yang, Yaodong and Fang, Cong and Li, Qing},
  booktitle=ICML,
  year={2024},
}

@inproceedings{delvecchio25nesysurvey,
  title     = {Neuro-Symbolic Artificial Intelligence: A Task-Directed Survey in the Black-Box Models Era},
  author    = {Delvecchio, Giovanni Pio and Molfetta, Lorenzo and Moro, Gianluca},
  booktitle = IJCAI,
  year      = {2025},
  note      = {Survey Track},
}

@article{Bhuyan24nesysurvey,
  author       = {Bikram Pratim Bhuyan and
                  Amar Ramdane{-}Cherif and
                  Ravi Tomar and
                  T. P. Singh},
  title        = {Neuro-symbolic artificial intelligence: a survey},
  journal      = {Neural Computing and Applications},
  volume       = {36},
  number       = {21},
  year         = {2024}
}

@inproceedings{Gupta22rewardshaping,
  author       = {Abhishek Gupta and
                  Aldo Pacchiano and
                  Yuexiang Zhai and
                  Sham M. Kakade and
                  Sergey Levine},
  title        = {Unpacking Reward Shaping: Understanding the Benefits of Reward Engineering
                  on Sample Complexity},
  booktitle    = NeuRIPS,
  year         = {2022}
}

@inproceedings{ng99shaping,
  author       = {Andrew Y. Ng and
                  Daishi Harada and
                  Stuart Russell},
  title        = {Policy Invariance Under Reward Transformations: Theory and Application
                  to Reward Shaping},
  booktitle    = ICML,
  year         = {1999}
}

@article{dreyfus1980five,
author = {Dreyfus, S.E. and Dreyfus, Hubert},
year = {1980},
title = {A Five-Stage Model of the Mental Activities Involved in Directed Skill Acquisition},
journal = {Distribution}
}

@book{bransford2000people,
  title={How People Learn: Brain, Mind, Experience, and School: Expanded Edition (2000)},
  author={Bransford, John D and Brown, Ann L and Cocking, Rodney R and others},
  volume={11},
  year={2000},
  publisher={Washington, DC: National Academy Press}
}

@inproceedings{TobinFRSZA17,
  author       = {Josh Tobin and
                  Rachel Fong and
                  Alex Ray and
                  Jonas Schneider and
                  Wojciech Zaremba and
                  Pieter Abbeel},
  title        = {Domain randomization for transferring deep neural networks from simulation to the real world},
  booktitle    = IROS,
  year         = {2017}
}

@inproceedings{PintoDSG17,
  author       = {Lerrel Pinto and
                  James Davidson and
                  Rahul Sukthankar and
                  Abhinav Gupta},
  title        = {Robust Adversarial Reinforcement Learning},
  booktitle    = ICML,
  year         = {2017}
}

@inproceedings{bacon2017option,
  title={The option-critic architecture},
  author={Bacon, Pierre-Luc and Harb, Jean and Precup, Doina},
  booktitle=AAAI,
  year={2017}
}

@inproceedings{bain1995framework,
  title={A Framework for Behavioural Cloning.},
  author={Bain, Michael and Sammut, Claude},
  booktitle={Machine Intelligence 15},
  year={1995}
}

@inproceedings{cheng2020,
  title={Policy improvement via imitation of multiple oracles},
  author={Cheng, Ching-An and Kolobov, Andrey and Agarwal, Alekh},
  booktitle=NeuRIPS,
  year={2020}
}

@InProceedings{nunes2003,
author="Nunes, Lu{\'i}s
and Oliveira, Eug{\'e}nio",
title="Exchanging Advice and Learning to Trust",
booktitle="Cooperative Information Agents VII",
series={Lecture Notes in Computer Science},
year="2003",
}

@article{ilhan2021student,
  title={Student-initiated action advising via advice novelty},
  author={Ilhan, Ercument and Gow, Jeremy and Perez, Diego},
  journal={{IEEE} Transactions on Games ({T-G})},
  year={2021},
  publisher={IEEE}
}

@inproceedings{farebrother2024cale,
  title={CALE: Continuous arcade learning environment},
  author={Farebrother, Jesse and Castro, Pablo Samuel},
  booktitle=NeuRIPS,
  year={2024}
}

@inproceedings{nexus,
title={Interpretable Reinforcement Learning via Meta-Policy Guidance},
author={Raban Emunds and Jannis Bl{\"u}ml and Quentin Delfosse and Kristian Kersting},
booktitle={The 18th European Workshop on Reinforcement Learning (EWRL)},
year={2025},
}

@inproceedings{gail,
  title={Generative adversarial imitation learning},
  author={Ho, Jonathan and Ermon, Stefano},
  booktitle=NeuRIPS,
  year={2016}
}

@article{ocatari,
  title={OCAtari: Object-Centric Atari 2600 Reinforcement Learning Environments},
  author={Delfosse, Quentin and Bl{\"u}ml, Jannis and Gregori, Bjarne and Sztwiertnia, Sebastian and Kersting, Kristian},
  year=2024,
  journal=RLJ
}

@article{huang2022cleanrl,
  title={Cleanrl: High-quality single-file implementations of deep reinforcement learning algorithms},
  author={Huang, Shengyi and Dossa, Rousslan Fernand Julien and Ye, Chang and Braga, Jeff and Chakraborty, Dipam and Mehta, Kinal and Ara{\~A}{\v{s}}jo, Jo{\~A}{\c{G}}o GM},
  journal=JMLR,
  year={2022}
}

@article{rajeswaran2017learning,
  title={Learning complex dexterous manipulation with deep reinforcement learning and demonstrations},
  author={Rajeswaran, Aravind and Kumar, Vikash and Gupta, Abhishek and Vezzani, Giulia and Schulman, John and Todorov, Emanuel and Levine, Sergey},
  journal={arXiv preprint arXiv:1709.10087},
  year={2017}
}

@article{huang2023guided,
  title={Guided reinforcement learning with efficient exploration for task automation of surgical robot},
  author={Huang, Tao and Chen, Kai and Li, Bin and Liu, Yun-Hui and Dou, Qi},
  journal={arXiv preprint arXiv:2302.09772},
  year={2023}
}

@inproceedings{bertoin2022look,
	title        = {Look Where You Look! Saliency-Guided Q-Networks for Generalization in Visual Reinforcement Learning},
	author       = {Bertoin, David and Zouitine, Adil and Zouitine, Mehdi and Rachelson, Emmanuel},
	year         = {2022},
	booktitle      = NeurIPS
}

@inproceedings{zhang2021learning,
	title        = {Learning Invariant Representations for Reinforcement Learning without Reconstruction},
	author       = {Amy Zhang and Rowan Thomas McAllister and Roberto Calandra and Yarin Gal and Sergey Levine},
	year         = {2021},
	booktitle    = ICLR
}

@inproceedings{yarats2021image,
	title        = {Image Augmentation Is All You Need: Regularizing Deep Reinforcement Learning from Pixels},
	author       = {Yarats, Denis and Kostrikov, Ilya and Fergus, Rob},
	year         = {2021},
	booktitle    = ICLR
}

@inproceedings{burda2019exploration,
  author       = {Yuri Burda and
                  Harrison Edwards and
                  Amos J. Storkey and
                  Oleg Klimov},
  title        = {Exploration by random network distillation},
  booktitle    = ICLR,
  year         = {2019},
}

@inproceedings{haarnoja2018soft,
  author       = {Tuomas Haarnoja and
                  Aurick Zhou and
                  Pieter Abbeel and
                  Sergey Levine},
  title        = {Soft Actor-Critic: Off-Policy Maximum Entropy Deep Reinforcement Learning
                  with a Stochastic Actor},
  booktitle    = ICML,
  year         = {2018},
}

@inproceedings{bellemare2016unifying,
  author       = {Marc G. Bellemare and
                  Sriram Srinivasan and
                  Georg Ostrovski and
                  Tom Schaul and
                  David Saxton and
                  R{\'{e}}mi Munos},
  title        = {Unifying Count-Based Exploration and Intrinsic Motivation},
  booktitle    = NeuRIPS,
  year         = 2016,
}

@inproceedings{c51,
  title={A distributional perspective on reinforcement learning},
  author={Bellemare, Marc G and Dabney, Will and Munos, R{\'e}mi},
  booktitle=ICLR,
  year={2017},
}

@article{sutton1999between,
	author = {Sutton, Richard S. and Precup, Doina and Singh, Satinder},
	journal = {Artificial Intelligence},
	number = {1-2},
	year = {1999},
	pages = {181--211},
	publisher = {},
	title = {Between {MDPs} and {Semi}-{MDPs}: A {Framework} for {Temporal} {Abstraction} in {Reinforcement} {Learning}.},
	volume = {112},
}

@article{dietterich2000hierarchical,
	author = {Dietterich, Thomas G.},
	journal = {Journal of Artificial Intelligence Research (JAIR)},
	year = {2000},
	pages = {227--303},
	publisher = {},
	title = {Hierarchical {Reinforcement} {Learning} with the {MAXQ} {Value} {Function} {Decomposition}.},
	volume = {13},
}

@inproceedings{vezhnevets2017feudal,
	author = {Vezhnevets, Alexander Sasha and Osindero, Simon and Schaul, Tom and Heess, Nicolas and Jaderberg, Max and Silver, David and Kavukcuoglu, Koray},
	booktitle = {International {Conference} on {Machine} {Learning} ({ICML})},
	year = {2017},
	pages = {3540--3549},
	organization = {},
	title = {FeUdal {Networks} for {Hierarchical} {Reinforcement} {Learning}.},
	volume = {},
}
\bibliographystyle{rlj}

\newpage
\appendix

\beginSupplementaryMaterials
\section*{Broader Impact Statement}
\label{sec:broaderImpact}
By integrating symbolic structure with deep reinforcement learning, \textit{Hybrid Hierarchical RL (\HHRL{})} seeks to improve alignment and long-horizon performance by discouraging short-term reward exploitation. However, 
badly designed or biased logical rules may cause the agent to rigidly follow flawed paths, producing behavior that is internally consistent yet unsatisfactory and difficult for neural refinement to override. 
\section{First-Order Logic}
\label{FOL}
In first-order logic, a term can be a constant, a variable, or a function term constructed using a function symbol. We denote an $n$-ary predicate ${\tt p}$ as ${\tt p}/(n, [{\tt dt_1}, \ldots, {\tt dt_n}])$, where ${\tt dt_i}$ represents the data type of the $i$-th argument.
An atom is an expression of the form ${\tt p(t_1, \ldots, t_n)}$, where ${\tt p}$ is an $n$-ary predicate symbol and ${\tt t_1}, \ldots, {\tt t_n}$ are terms. If the atom contains no variables, it is referred to as a ground atom, or simply a fact.

A literal is either an atom or the negation of an atom. We refer to an atom as a positive literal, and its negation as a negative literal.
A clause is defined as a finite disjunction ($\lor$) of literals. When a clause contains no variables, it is called a ground clause.
A definite clause is a special case: a clause that contains exactly one positive literal. Formally, if $A, B_1, \ldots, B_n$ are atoms, then the expression $A \lor \lnot B_1 \lor \ldots \lor \lnot B_n$ constitutes a definite clause. We write definite clauses in the form of $A~\mbox{:-}~B_1,\ldots, B_n$. where $A$ is the {\it head} of the clause, and the set $\{B_1, \ldots, B_n\}$ is referred to as the body. For simplicity, we refer to definite clauses as clauses throughout this paper. The forward-chaining inference is a type of inference in first-order logic to compute logical entailment~\citep{Russel09}.


\section{Architecture of the Hierarchical DQN (hDQN)}
\label{app:hdqn-arch}

We describe here the hierarchical deep Q-network (hDQN) custom implementation. The agent follows the standard two-level hDQN structure: a \emph{meta-controller} that selects a temporally-extended \emph{subgoal} $g$, and a \emph{controller} that selects primitive actions $a$ conditioned on both the environment state and the current goal.

\subsection{Meta-Controller Network}
The meta-controller estimates the goal-values
\begin{equation}
Q_{\theta_m}(s,g) \in \mathbb{R}^{G},
\end{equation}
where $G$ is the number of subgoals (varies by game:  $G_\text{Seaquest} = 8$,  $G_\text{Kangaroo} = 6$,  $G_\text{DonkeyKong} = 10$).

\paragraph{Input/Output.}
\begin{itemize}
    \item Input: stacked grayscale frames $s \in \mathbb{R}^{4 \times 84 \times 84}$ (the network expects 4 channels).
    \item Output: goal Q-values $Q_{\theta_m}(s,\cdot)\in\mathbb{R}^{G}$.
    \item Normalization: the forward pass normalizes the 8-bit gray-scale input image to a $[0, 1]^{84 \times 84}$ float array.
\end{itemize}

\paragraph{Architecture.}
The meta-controller is a DQN-style CNN. The architecture is shown in Tab.~\ref{tab:meta_controller_arch}

\begin{table}[h!]
    \centering
    \caption{Meta-Controller Architecture}
    \begin{tabular}{cc}
        \toprule
        Layer           & Configuration                       \\
        \midrule
        Convolutional   & Conv2d(4, 32, 8, stride=4)          \\
                        & ReLU                                \\
        Convolutional   & Conv2d(32, 64, 4, stride=2)         \\
                        & ReLU                                \\
        Convolutional   & Conv2d(64, 64, 3, stride=1)         \\
                        & ReLU                                \\
                        & Flatten                             \\
        Fully Connected & Linear(64 $\cdot$ 7 $\cdot$ 7, 512) \\
                        & ReLU                                \\
        Fully Connected & Linear(512, $G$)                    \\
        \bottomrule
    \end{tabular}
    \label{tab:meta_controller_arch}
\end{table}

\subsection{Controller Network}
The controller estimates action-values conditioned on the goal:
\begin{equation}
Q_{\theta_c}(s,g,a) \in \mathbb{R}^{|\mathcal{A}|}.
\end{equation}

\paragraph{Input/Output.}
\begin{itemize}
    \item Input: a goal-conditioned image tensor in $\mathbb{R}^{5 \times 84 \times 84}$ (the network expects 5 channels).
    \item Output: action Q-values $Q_{\theta_c}(s,g,\cdot)\in\mathbb{R}^{|\mathcal{A}|}$, where $|\mathcal{A}|=$ number of actions.
\end{itemize}

\paragraph{Architecture.}
The controller matches the meta-controller CNN except for the first convolution input channels (cf. Tab.~\ref{tab:controller_arch}).
\begin{table}[h!]
    \centering
    \caption{Controller Network Architecture}
    \begin{tabular}{cc}
        \toprule
        Layer           & Configuration                       \\
        \midrule
        Convolutional   & Conv2d(5, 32, 8, stride=4)          \\
                        & ReLU                                \\
        Convolutional   & Conv2d(32, 64, 4, stride=2)         \\
                        & ReLU                                \\
        Convolutional   & Conv2d(64, 64, 3, stride=1)         \\
                        & ReLU                                \\
                        & Flatten                             \\
        Fully Connected & Linear(64 $\cdot$ 7 $\cdot$ 7, 512) \\
                        & ReLU                                \\
        Fully Connected & Linear(512, $|\mathcal{A}|$)        \\
        \bottomrule
    \end{tabular}
    \label{tab:controller_arch}
\end{table}

\section{Visual Encoder Architecture of exPPO.}
\label{app:exppoarch}
In this section, we introduce the visual encoder architecture of exPPO. The Actor, Critic architecture stays the same with PPO (See Tabs.~\ref{tab:exppo_pixel_encoder} and \ref{tab:exppo_obj_encoder}). $N$ is an environment-dependent constant that fixes the flattened size. In our implementation, $N_\text{Kangaroo} = 49$, yielding a linear layer with input size of 12,544. $N_\text{Seaquest} = 43$, and $N_\text{DonkeyKong} = 18$.


\begin{table}[h!]
    \centering
    \caption{Visual Encoder Architecture (Pixel Stream)}
    \begin{tabular}{cc}
        \toprule
        Layer           & Configuration                       \\
        \midrule
        Convolutional   & Conv2d(4, 32, 8, stride=4)          \\
                        & ReLU                                \\
        Convolutional   & Conv2d(32, 64, 4, stride=2)         \\
                        & ReLU                                \\
        Convolutional   & Conv2d(64, 64, 3, stride=1)         \\
                        & ReLU                                \\
                        & Flatten                             \\
        Fully Connected & Linear(64 $\cdot$ 7 $\cdot$ 7, 512) \\
                        & ReLU                                \\
        \bottomrule
    \end{tabular}
    \label{tab:exppo_pixel_encoder}
\end{table}


\begin{table}[h!]
    \centering
    \caption{Object Encoder Architecture (Symbolic Stream)}
    \begin{tabular}{cc}
        \toprule
        Layer           & Configuration                      \\
        \midrule
        Fully Connected & Linear(4, 128)                     \\
                        & ReLU                               \\
        Fully Connected & Linear(128, 64)                    \\
                        & ReLU                               \\
                        & Flatten                            \\
        Fully Connected & Linear(64 $\cdot$ 4 $\cdot$ $N$, 32) \\
                        & ReLU                               \\
        \bottomrule
    \end{tabular}
    \label{tab:exppo_obj_encoder}
\end{table}

\subsection{Fusion Trunk (Shared Representation)}
The model concatenates the two modality embeddings and maps them to a shared 32-dimensional hidden state:

\begin{table}[h!]
\centering
\caption{Fusion Trunk Architecture (Pixel + Symbolic Fusion)}
\label{tab:fusion_trunk_arch}
\begin{tabular}{l l}
\toprule
\text{Layer} & \text{Configuration} \\
\midrule
Concatenation & Concat($y^{\text{pixel}}\!\in\!\mathbb{R}^{512}$, $y^{\text{symb.}}\!\in\!\mathbb{R}^{32}$) \\
Fully Connected & Linear(544, 32) \\
 & ReLU \\
\bottomrule
\end{tabular}
\end{table}

\section{Training Parameters}
The options and models were all trained using the same hyperparameters listed in Tab.~\ref{tab:hyper} because of time- and hardware constraints. The models were all trained on an Nvidia RTX 2090 with 12 GB.
\begin{table}[t]
    \centering
    \caption{Hyperparameters used when training the models.}
    \begin{tabular}{cc}\toprule
        Parameter & Value \\\midrule
        Gamma & 0.99\\ 
        Learning rate & 0.00025 \\
        Blending entropy coefficient & 0.01 \\
        Entropy coefficient & 0.01 \\
        Clipping coefficient & 0.1 \\
        Maximum gradient norm &  0.5\\
        Number of environments & 128 \\
        Number of steps & 128 \\
        Total timesteps & 40,000,000\\ \bottomrule
    \end{tabular}
    \label{tab:hyper}
\end{table}
In our experiments, we use the CleanRL library \citep{huang2022cleanrl} for the PPO, C51, and DQN implementations. All neural networks that use the pixel state have the same layer configuration described in Tab \ref{tab:layers_pixel_state}. Furthermore, a simple linear layer is added to the critic or the actor, depending on the algorithm. For the object-centric critic, the layer configuration is described in Tab \ref{tab:layers_object_centric}.

\begin{table}[ht]
    \centering
    \caption{Pixel State Backbone Architecture}
    \begin{tabular}{cc}
        \toprule
        Layer           & Configuration                       \\
        \midrule
        Convolutional   & Conv2d(4, 32, 8, stride=4)          \\
                        & ReLU                                \\
        Convolutional   & Conv2d(32, 64, 8, stride=2)         \\
                        & ReLU                                \\
        Convolutional   & Conv2d(64, 64, 3, stride=1)         \\
                        & ReLU                                \\
                        & Flatten                             \\
        Fully Connected & Linear(64 $\cdot$ 7 $\cdot$ 7, 512) \\
                        & ReLU                                \\
        \bottomrule
     \end{tabular}
    \label{tab:layers_pixel_state}
\end{table}

\begin{table}[ht]
    \centering
    \caption{Object-Centric Backbone Architecture}
    \begin{tabular}{cc}\toprule
         Layer&  Configuration\\\midrule
         Fully Connected& 
    Linear($N_{in}$, 512)\\
 &ReLU\\
 Fully Connected&Linear(512, 120)\\
 &ReLU\\
 Fully Connected&Linear(120, $N_{out}$)\\ \bottomrule\end{tabular}
    \label{tab:layers_object_centric}
\end{table}
\section{Making Symbolic Reasoning Differentiable}
\label{diff}
We now describe each step in detail for making the symbolic planner differentiable.

\paragraph{(Step 1) Encoding Logic Programs as Tensors.}
To enable differentiable forward reasoning, each meta-rule is converted to a tensor representation. Each meta-rule $C_i \in \mathcal{C}$ is encoded as a tensor $\mathbf{I}_i \in \mathbb{N}^{G \times S \times L}$, where $S$ denotes the maximum number of substitutions for existentially quantified variables in the rule set, and $L$ is the maximum number of atoms in the body of any rule. 
For instance, $\mathbf{I}_i[j, k, l]$ stores the index of the $l$-th subgoal in the body of rule $C_i$ used to derive the $j$-th fact under the $k$-th substitution.

\paragraph{(Step 2) Weighting and Selecting Meta-Rules.}
We construct the reasoning function by assigning weights that determine how multiple meta-rules are combined.
(i) We fix the size of the target meta-program to be $M$, meaning the final program will consist of $M$ meta-rules selected from a total of $C$ candidates in $\mathcal{C}$.
(ii) To enable soft selection, we define a weight matrix $\mathbf{W} = [\mathbf{w}_1, \ldots, \mathbf{w}_M]$, where each $\mathbf{w}_i \in \mathbb{R}^C$ assigns a real-valued weight.
(iii) We then apply a \emph{softmax} to each $\mathbf{w}_i$ to obtain a probability distribution over the $C$ candidates, allowing the model to softly combine multiple meta-rules. 
\paragraph{(Step 3) Perform Differentiable Inference.}
Starting from a single application of the weighted meta-rules, we iteratively propagate inferred facts across $T$ reasoning steps. 

We compute the valuation of body atoms for every grounded instance of a meta-rule $C_i \in \mathcal{C}$. This is achieved by first gathering the current truth values from the valuation vector $\mathbf{v}^{(t)}$ using an index tensor $\mathbf{I}i$, and then applying a multiplicative aggregation across subgoals:
\begin{align}
b_{i,j,k}^{(t)} = \prod_{l=1}^{L} \mathbf{gather}(\mathbf{v}^{(t)}, \mathbf{I}i)[j,k,l],
\label{eq:gather_prod_body}
\end{align}
where the $\mathbf{gather}$ operator maps valuation scores to indexed body atoms:
\begin{align}
\mathbf{gather}(\mathbf{x}, \mathbf{Y})[j,k,l] = \mathbf{x}[\mathbf{Y}[j,k,l]].
\end{align}
The resulting value $b_{i,j,k}^{(t)} \in [0,1]$ reflects the conjunction of subgoal valuations under the $k$-th substitution of existential variables, used to derive the $j$-th candidate fact from the $i$-th meta-rule. Logical conjunction is implemented via element-wise product, modeling the "and" over the rule body.

To integrate the effects of multiple groundings of a meta-rule $C_i$, we apply a smooth approximation of logical \emph{or} across all possible substitutions. Specifically, we compute the aggregated valuation $c^{(t)}_{i,j} \in [0,1]$ as:
\begin{align}
c^{(t)}_{i,j} = \mathit{softor}^\gamma(b_{i,j,1}^{(t)}, \ldots, b_{i,j,S}^{(t)}),
\end{align}
where $\mathit{softor}^\gamma$ denotes a differentiable relaxation of disjunction. This operator is defined as:
\begin{align}
\mathit{softor}^\gamma(x_1, \ldots, x_n) = \gamma \log \sum_{i=1}^{n} \exp(x_i / \gamma),
\label{eq:softor}
\end{align}
with temperature parameter $\gamma > 0$ controlling the smoothness of the approximation. This formulation closely resembles a softmax over valuations and serves as a continuous surrogate for the logical \emph{max}, following the log-sum-exp technique commonly used in differentiable reasoning \citep{cuturi2017soft}.

\textbf{(ii) Weighted Aggregation Across Meta-Rules.}
We compute a weighted combination of meta rules using the learned soft selections:
\begin{align}
h_{j,m}^{(t)} = \sum_{i=1}^{C} w^{m,i} \cdot c^{(t)}{i,j},
\end{align}
where $h^{(t)}_{j,m} \in [0,1]$ represents the intermediate result for the $j$-th fact contributed by the $m$-th slot. Here, $w^{m,i}$ is the softmax-normalized score over the $i$-th meta-rule:
\begin{align*}
w^*_{m,i} = \frac{\exp(w_{m,i})}{\sum_{i'} \exp(w_{m,i'})}, \quad w_{m,i} = \mathbf{w}_m[i].
\end{align*}

Finally, we consolidate the outputs of the $M$ softly selected rule components using a smooth disjunction:
\begin{align}
r_j^{(t)} = \mathit{softor}^\gamma(h^{(t)}_{j,1}, \ldots, h^{(t)}_{j,M}),
\label{eq:softor_weighted_rules}
\end{align}
which yields the $t$-step valuation for fact $j$. This mechanism allows the model to integrate $M$ soft rule compositions from a larger pool of $C$ candidates in a fully differentiable way.


\textbf{(iii) Iterative Forward Reasoning.}
We iteratively apply the forward reasoning procedure for $T$ steps. At each step $t$, we update the valuation of each fact $j$ by softly merging its newly inferred value $r_j^{(t)}$ with its previous valuation:
\begin{align}
v^{(t+1)}_j = \mathit{softor}^\gamma(r_j^{(t)}, v_j^{(t)}).
\label{eq:softor_iterative_FR}
\end{align}
This recursive update mechanism approximates logical entailment in a differentiable form, enabling the model to perform $T$-step reasoning over the evolving fact valuations.
The whole reasoning computation Eqs.~\ref{eq:gather_prod_body}--\ref{eq:softor_iterative_FR} can be implemented using efficient tensor operations.

\section{Reward Comparison}
\label{app.reward_comparison}
We provide a comparison of rewards across different methods pre-trained with the \HHRL{} architecture and its counterpart in Tab.~\ref{tab:rq2table}.

\begin{table*}[t]
\small
\centering
\caption{\textbf{\HHRL{} outperforms other compared baselines in the Continuous Atari learning environment (CALE).} 
Reward comparison of \HHRL{} against other baseline methods which can work in the continuous action space versions of Kangaroo and DonkeyKong.}

\begin{tabular}{lcccccccc}
\toprule
                    & \HHRL{} & PPO & hPPO & hReason \\ 
\midrule
Kangaroo (cont.)    & 84665\spm{49767} & 1785\spm{72} & 19854\spm{18586}& 557\spm{167} \\
DonkeyKong (cont.)  & 10818\spm{7431} & 3836\spm{530} &991.0\spm{446} & 542\spm{975} \\
\bottomrule
\end{tabular}
\label{tab:rq5table}
\end{table*}


\begin{table*}[ht!]
\small
\centering
\caption{\textbf{\HHRL{} outperforms other compared baselines in the classical Atari Learning environment.} Although PPO and hPPO achieve relatively high scores in Kangaroo, their improvements stem from policy misalignment. We evaluate \HHRL{} against other methods in Kangaroo, Seaquest, and DonkeyKong. $^\ast$Due to instability during evaluation, we report the maximum performance achieved by hDQN during training as a representative metric.}

\centering
\begin{tabular}{lccccc}
\toprule
           & \hhrlpp       & \hhrlp           & \HHRL{}        & PPO           & NUDGE       \\
\midrule
Seaquest   & 4759\spm{1004}   & 1802\spm{400}    & 2812\spm{1477} &3247\spm{881} & 63\spm{18}  \\
Kangaroo   & 131842\spm{1221} & 2754\spm{2626}   & 5148\spm{3950}    & 14592\spm{491}   & 404\spm{230}  \\
DonkeyKong & 216793\spm{125655} & 87780\spm{32786} & 33690\spm{14565}   & 4536\spm{296}   & 21\spm{43}   \\
\bottomrule
           &               &                  &                &               &             \\
\toprule
           & BlendRL       & Option-critic    & hDQN           & hPPO          & hReason     \\
\midrule
Seaquest   & 117\spm{62}   & 38\spm{19}        & 112$^\ast$     & 1906\spm{628}   & 1186\spm{930} \\
Kangaroo   & 1482\spm{1343}   & 61\spm{89}        & 322$^\ast$     & 10601\spm{915}  & 1701\spm{1061} \\
DonkeyKong & 85\spm{82}     & ---              & 434$^\ast$     & 418\spm{139}    & 860\spm{1301}  \\
\bottomrule
\end{tabular}
\label{tab:rq1table}
\end{table*}

\section{Options pretraining}
\label{app.options_pretrain}
The options for the HHRL framework were created by training multiple PPO agents in modified Atari Environments, each with its own custom reward function. The reward function, modifications, and the number of options were obtained from domain experts. More details about the modification can be found in Tab.~\ref {tab:mods}, which modification was used to train certain options in Tab.~\ref {tab:option_mods}, and the reward functions in Tab.~\ref {tab:option_rewards}.

\begin{table}[t]
    \centering
    \setlength{\tabcolsep}{2pt} 
    \renewcommand{\arraystretch}{1.05} 
    \caption{A list of modifications for each environment.}
    \begin{tabular}{llp{7.25cm}}
    \toprule
         Environment & Modification&  Description\\\midrule
         Seaquest&random\_start& 
    randomize the starting position of the player\\
 &random\_surfacing&set the player to a random position when reaching the surface\\
 &random\_lane&set the starting position of the player  in one of the 4 lanes where enemies and divers spawn\\
 &disable\_underwater\_enemies& prevent all underwater enemies \newline from appearing in the game\\
 &disable\_enemies& prevent all enemies from appearing in the game\\
 &unlimited\_collected\_diver& always set the amount of collected divers to 1\\
 &empty\_divers\_automatically& reset the amount of collected divers to 0 if 5 or more divers were collected\\
 &enable\_surface\_enemies&enable the surface submarine\\
 &unlimited\_oxygen&prevent the oxygen bar from decreasing\\
 \midrule
 Kangaroo&disable\_coconut& 
    keep falling coconuts on the top of the screen\\
 &disable\_thrown\_coconut&remove coconuts thrown by the monkeys from the game\\
 &disable\_high\_thrown\_coconut&thrown coconuts are always thrown along the ground \\
 &disable\_low\_thrown\_coconut&thrown coconuts are always thrown in the air\\
 &disable\_monkeys&remove enemy monkeys from the game \\
 &plattform\_checkpoints&place the player on the last plattform they stood on after losing a life\\
 &randomize\_kangaroo\_position&randomize the starting position of the player\\
 &teleport\_kill&set the player to a random position after defeating a monkey\\
 &always\_falling\_coconut&directly spawn another falling coconut above the player after another one has already fallen below the player\\
 &change\_level\_0&always set the current level to 0\\\midrule
 Donkey Kong&skip\_start\_game& 
    remove the need to press space to start the game\\
 &spawn\_barrel\_in\_random\_lane&randomize the starting position of barrels\\
 &remove\_near\_barrel&remove a barrel if the player is near it\\
 &spawn\_barrel\_near\_player&place one barrel in front of the player at the start and after the player has jumped over it\\
 &no\_hammer&remove the hammer from the game\\
 &shorten\_timelimit&shorten the timelimit to reach the princess to half the original amount\\
 &random\_start&randomize the starting position of the player\\
 &end\_hammer&end the episode after the hammer has been used\\
 &spawn\_near\_hammer&set the starting position of the player randomly near the hammer position\\
 &change\_level\_0&always set the current level to 0\\\midrule
 Bankheist&random\_city\_start& 
    randomize the city the player starts in\\
 &random\_start&randomize the starting position of the player\\
 &two\_police\_cars&always have two police cars and one bank on the field\\
 &delete\_near\_police&remove a police car if it is near the player\\
 \bottomrule\end{tabular}
    \label{tab:mods}
    
\end{table}
\begin{table}[ht!]
    \centering
    \renewcommand{\arraystretch}{1.1} 
    \caption{A list of modifications used when training options in each environment.}
    \label{tab:option_mods}
    \begin{tabular}{ l l l }
    \toprule
         Environment & Option&  Applied modifications\\\midrule
         Seaquest& get\_air, deliver\_diver & random\_start \\
                         &&disable\_underwater\_enemies \\
                         &&unlimited\_collected\_diver\\
                         &&empty\_divers\_automatically\\
                         &&random\_surfacing\\
                         &&enable\_surface\_enemies\\
                & get\_diver & random\_start \\
                                &&unlimited\_oxygen \\
                                &&emtpy\_divers\_automatically\\
                & shoot\_enemy/wait\_in\_line &random\_lane\\
                &&unlimited\_oxygen\\\midrule
 Kangaroo& ascend & disable\_coconut \\
                 &&disable\_thrown\_coconut\\
                 &&plattform\_checkpoints\\
                 &&change\_level\_0\\
        & deal\_with\_enemies & randomize\_kangaroo\_position\\
                 &&disable\_coconut\\
                 &&teleport\_kill\\
                 &&change\_level\_0\\
        & avoid\_thrown\_coconuts & randomize\_kangaroo\_position\\
             &&disable\_coconut\\
            &&change\_level\_0\\
        & avoid\_low\_thrown\_coconuts & randomize\_kangaroo\_position\\
             &&disable\_coconut\\
             &&disable\_high\_thrown\_coconut\\
            &&change\_level\_0\\
        & avoid\_high\_thrown\_coconuts &randomize\_kangaroo\_position\\
             &&disable\_coconut\\
             &&disable\_low\_thrown\_coconut\\
            &&change\_level\_0
        \\\midrule
 Donkey Kong& climb & remove\_near\_barrel\\
                 &&skip\_start\_game\\
                 &&change\_level\_0\\
        & use\_hammer & random\_start\\
                 &&spawn\_barrel\_near\_player\\
                 &&no\_hammer\\
                 &&shorten\_timelimit\\
                 &&skip\_start\_game\\
                 &&change\_level\_0\\
        & jump\_barrel & end\_hammer\\
                     &&spawn\_near\_hammer\\
                     &&skip\_start\_game\\
                     &&change\_level\_0\\
 \bottomrule\end{tabular}
\end{table}

\begin{table}[ht!]
    \centering
    \setlength{\tabcolsep}{3pt} 
    \renewcommand{\arraystretch}{1.1} 
    \caption{A list of reward functions used when training options in each environment.}
    \begin{tabular}{ l p{4.2cm} p{7cm} }
    \toprule
         Environment & Option&  Reward function\\\midrule
         Seaquest& get\_air, deliver\_diver & +1.5 when player is at the surface\\
                & get\_diver & +1 when player collects a diver\\
                & shoot\_enemy/wait\_in\_line & +1 when the player shoots an enemy\\ 
                && -10 when the player moves up or down\\\midrule
 Kangaroo& ascend & +20 when player reaches certain positions\\
        & deal\_with\_enemies & +1 when the player obtains a score from punching an enemy\\
        & & -1 when the player loses a life\\
        & avoid\_thrown\_coconuts/ avoid\_high\_thrown\_coconuts  & +1 once when a thrown coconut is on the left side of the player \\
         && -1 reward when the player moves left or right\\
         &  avoid\_low\_thrown\_coconuts & +1 when player has jumped over a thrown coconut \\
         && -1 when the player moves left or right\\\midrule
 Donkey Kong& climb & +20 when player reaches certain positions \\
        & use\_hammer & 10 when player grabs the hammer\\
             && +1 when player obtains score points with the hammer\\
        & jump\_barrel & +1 when player jumped over a barrel\\
 \bottomrule\end{tabular}
    \label{tab:option_rewards}
\end{table}

\clearpage

\section{Logic Rules}
\label{app:logicrules}
\textbf{Seaquest} \\
Logic manager rules:
\begin{small}
\begin{verbatim}
get_air(X)       :- oxygen_low(B), collected_at_least_one_diver(X),
                    not_below_enemy(X).
deliver_diver(X) :- full_divers(X),
                    oxygen_high(B).
shoot_enemy(X)   :- same_depth_enemy(P,E), same_depth_diver(P,D),
                    not_full_divers(X).
wait_in_lane(X)  :- below_enemy(X), not_same_depth_diver(X).
wait_in_lane(X)  :- above_enemy(X), not_same_depth_diver(X), oxygen_high(B). 
wait_in_lane(X)  :- no_visible_divers(D), oxygen_high(B), not_full_divers(X). 
get_diver(X)     :- visible_diver(D), not_same_depth_enemy(X),
                    not_full_divers(X), oxygen_high(B).
\end{verbatim}
\end{small}
Logic gating rule:
\begin{small}
\begin{verbatim}
logic_agent(X)  :- oxygen_low(B).
logic_agent(X)  :- full_divers(X).
logic_agent(X)  :- visible_diver(D).
neural_agent(X) :- oxygen_high(B), no_visible_divers(X), not_full_divers(X).
\end{verbatim}
\end{small}
\textbf{Kangaroo}\\
Logic manager rules (discrete):
\begin{small}
\begin{verbatim}
ascend(X)                :- nothing_around(X). 
ascend(X)                :- on_ladder(P,L).
deal_with_enemy(X)       :- close_by_monkey(P,M),
                            not_below_fallingcoconut(P,FC).
avoid_thrown_coconuts(X) :- close_by_throwncoconut(P,TC), not_on_ladder(X).
\end{verbatim}
\end{small}
Logic manager rules (continuous):
\begin{small}
\begin{verbatim}
ascend(X)                     :- nothing_around(X). 
ascend(X)                     :- on_ladder(P,L).
deal_with_enemy(X)            :- close_by_monkey(P,M),
                                 not_below_fallingcoconut(P,FC).
avoid_low_thrown_coconuts(X)  :- close_by_low_throwncoconut(P,TC),
                                 not_on_ladder(X). 
avoid_high_thrown_coconuts(X) :- close_by_high_throwncoconut(P,TC),
                                 not_on_ladder(X).
\end{verbatim}
\end{small}
Logic gating rule:
\begin{small}
\begin{verbatim}
logic_agent(X)  :- low_on_time(T).
logic_agent(X)  :- nothing_around(X).
neural_agent(X) :- anything_around(X), plenty_time(T).
\end{verbatim}
\end{small}
\textbf{Donkey Kong}\\
Logic manager rules:
\begin{small}
\begin{verbatim}
climb(X)       :- nothing_around(X).
use_hammer(X)  :- hammer_on_level(P,H).
jump_barrel(X) :- close_by_barrel(P,B).
\end{verbatim}
\end{small}
Logic gating rule:
\begin{small}
\begin{verbatim}
logic_agent(X)  :- hammer_on_level(P,H).
logic_agent(X)  :- nothing_around(X).
neural_agent(X) :- close_by_barrel(P,B).
\end{verbatim}
\end{small}




%
%

\end{document}